\documentclass[runningheads]{llncs}
\usepackage[T1]{fontenc}
\usepackage{times}
\usepackage{epsfig}
\usepackage{graphicx}
\usepackage{amsmath}
\usepackage{amssymb}
\usepackage{array}
\usepackage{multirow}
\usepackage{subcaption}
\usepackage{color}
\usepackage{float}
\usepackage{graphicx}
\usepackage{mathtools}
\usepackage{booktabs}
\usepackage{hyperref}
\usepackage{fancyhdr}
\fancypagestyle{specialfooter}{%
  \fancyhf{}
  
  \fancyhead[L]{\scriptsize{Accepted in International Conference on Computer Vision and Image Processing (CVIP), 2022}}
  \fancyfoot[L]{\scriptsize{$^3$Work done while at IIIT Sri City}}
}

\begin{document}
\title{LiSHT: Non-Parametric Linearly Scaled Hyperbolic Tangent Activation Function for Neural Networks}
\titlerunning{LiSHT Activation Function}

\author{Swalpa Kumar Roy\inst{1} \and
Suvojit Manna\inst{2} \and
Shiv Ram Dubey\inst{3} \and Bidyut Baran Chaudhuri\inst{4}}

\authorrunning{Roy et al.}

\institute{Department of Computer Science \& Engineering, Jalpaiguri Government of Engineering College, West Bengal, India \and
CureSkin, Bengaluru, Karnataka, India \and
Computer Vision and Biometrics Lab, Indian Institute of Information Technology, Allahabad, Uttar Pradesh, India \and Techno India University, Kolkata, India and Indian Statistical Institute, Kolkata, India \\
\email{swalpa@cse.jgec.ac.in, suvojit@heallo.ai, srdubey@iiita.ac.in, bidyutbaranchaudhuri@gmail.com}}

\maketitle
\thispagestyle{specialfooter}

\begin{abstract}
The activation function in neural network introduces the non-linearity required to deal with the complex tasks. Several activation/non-linearity functions are developed for deep learning models. However, most of the existing activation functions suffer due to the dying gradient problem and non-utilization of the large negative input values. In this paper, we propose a Linearly Scaled Hyperbolic Tangent (LiSHT) for Neural Networks (NNs) by scaling the Tanh linearly. The proposed LiSHT is non-parametric and tackles the dying gradient problem. We perform the experiments on benchmark datasets of different type, such as vector data, image data and natural language data. We observe the superior performance using Multi-layer Perceptron (MLP), Residual Network (ResNet) and Long-short term memory (LSTM) for data classification, image classification and tweets classification tasks, respectively. The accuracy on CIFAR100 dataset using ResNet model with LiSHT is improved by 9.48, 3.40, 3.16, 4.26, and 1.17\% as compared to Tanh, ReLU, PReLU, LReLU, and Swish, respectively. We also show the qualitative results using loss landscape, weight distribution and activations maps in support of the proposed activation function.

\keywords{Activation Function \and Convolutional Neural Networks \and Non-Linearity \and Tanh function \and Image Classification.}
\end{abstract}

\section{Introduction}
The deep learning method is one of the breakthroughs which replaced the hand-tuning tasks in many problems including computer vision, speech processing, natural language processing, robotics, and many more \cite{schmidhuber2015deep}, \cite{garcia2017review}, \cite{voulodimos2018deep}, \cite{dubey2021decade}, \cite{dubey2022vision}. In recent times, the deep Artificial Neural Networks (ANNs) have shown a tremendous performance improvement due to existence of larger datasets as well as powerful computers \cite{goodfellow2016deep}. 
Various types of ANN have been proposed for several problems such as Multilayer Perceptron (MLP)~\cite{hornik1989multilayer} to deal with the real vector $R$-dimensional data \cite{kim2018deepx}. Convolutional Neural Networks (CNN) are used to deal with the image and videos \cite{resnet}. Recurrent Neural Network (RNN) like Long-Short Term Memory (LSTM) are used for the sentiment analysis \cite{wang2016attention}.
The main aim of different type of neural networks is to transform the input data in abstract feature space. In order to achieve it, all the neural networks rely on a compulsory unit called the activation function~\cite{agostinelli2014learning}. The activation functions bring the non-linear capacity in the network to deal with the complex data \cite{dubey2022activation}.

The $Sigmoid$ activation function was mostly used in the at the inception of neural networks. It is a special case of the logistic function. The $Sigmoid$ function squashes the real-valued numbers into $0$ or $1$. In turn, the large negative number becomes $0$ and large positive number becomes $1$. The hyperbolic tangent function $Tanh$ is the another popular activation function. The output range of $Tanh$ is defined with $-1$ as lower limit and $1$ as upper limit. The vanishing gradient in both positive as well as negative directions is one of the major problems with both $Sigmoid$ and $Tanh$ activation functions. The Rectified Linear Unit ($ReLU$) activation function was proposed in recent past for training deep networks \cite{alexnet}. $ReLU$ is a breakthrough against vanishing gradient. It is a zero function (i.e., the output is zero) for the negative inputs and an identity function for the positive inputs. The $ReLU$ is very simple, hence became very popular and mostly used in different deep models. The diminishing gradient for the inputs less than zero can be seen as primary bottleneck with $ReLU$ leading to dying gradient problem.

Several researchers have proposed the improvement on $ReLU$ such as Leaky ReLU ($LReLU$) \cite{lrelu}, Parametric ReLU ($PReLU$) \cite{prelu}, $Softplus$ \cite{nair2010rectified}, Exponential Linear Unit ($ELU$) \cite{clevert2015fast}, Scaled Exponential Linear Unit ($SELU$) \cite{klambauer2017self}, Gaussian Error Linear Unit ($GELU$) \cite{hendrycks2016bridging}, Average Biased ReLU ($ABReLU$) \cite{abrelu}, Linearized sigmoidal activation ($LiSA$)~\cite{bawa2019linearized} etc. 
The $ReLU$ is extended to $LReLU$ by allowing a small, non-negative and constant gradient (such as 0.01) for the negative inputs \cite{lrelu}. The $PReLU$ makes the slopes of linear function for negative inputs (i.e., leaky factor) as trainable \cite{prelu}. The $Softplus$ activation function tries to make the transition of $ReLU$ (i.e., at $0$) smooth by fitting the log function \cite{nair2010rectified}. Otherwise, the $Softplus$ activation is very similar to the $ReLU$ activation. The $ELU$ function is same as $ReLU$ for positive inputs and exponential for negative inputs \cite{clevert2015fast}. The $ELU$ becomes smoother near zero. For positive inputs, the $ELU$ \cite{clevert2015fast} can blow up the activation, which can lead to the gradient exploding problem. The $SELU$ adds one scaling parameter in $ELU$, which makes it better against weight initialization \cite{klambauer2017self}. The $GELU$ uses a Gaussian approach to apply the zero/identity map to the input of a unit randomly \cite{hendrycks2016bridging}. The $ABReLU$ utilizes the representative negative values as well as representative positive values by shifting rectification based on the average of activation values \cite{abrelu}. The $ABReLU$ also could not utilize all the negative values due to trimming of values at zero, similar to $ReLU$. Most of these existing activation methods are sometimes not able to take the advantage of negative values which is solved in the proposed $LiSHT$ activation.

Recently, Xu et al. have performed an empirical study of rectified activations in CNNs \cite{xu2015empirical}. Very recently, a promising $Swish$ activation function was introduced as sigmoid-weighted linear unit, i.e., $f(x) = x \times sigmoid(\beta x)$ \cite{ramachandran2017swish}. Based on the value of the learnable $\beta$, $Swish$ adjusts the amount of non-linearity. 

In this paper, a linearly scaled hyperbolic tangent activation function ($LiSHT$) is proposed to introduce the non-linearities in the neural networks. The $LiSHT$ scales the $Tanh$ function linearly to tackle its gradient diminishing problem. 

\begin{figure*}[!t]
\centering
\begin{tabular}{ccc}
\includegraphics[clip=true, trim = 135 285 120 280, width=.30\linewidth]{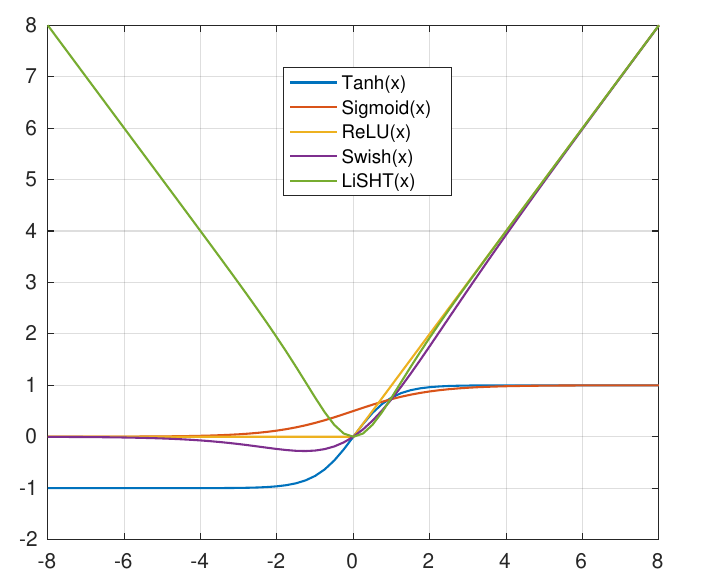} 
&
\includegraphics[clip=true, trim = 135 290 120 280, width=.32\linewidth]{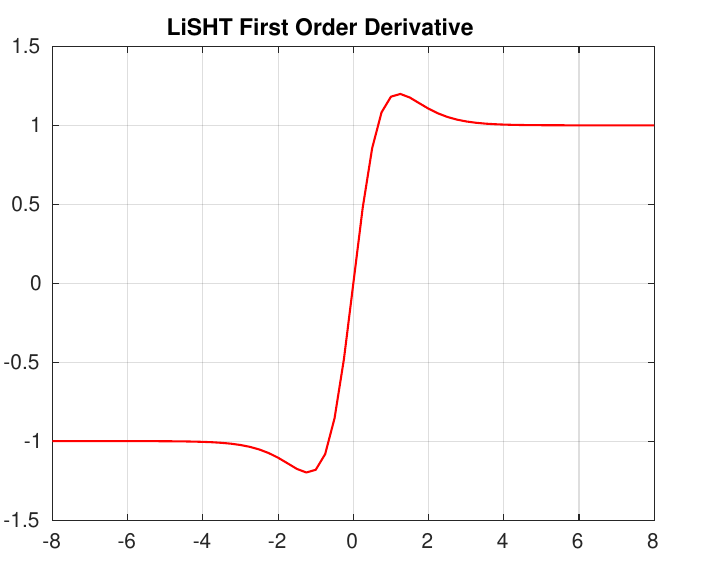} 
&
\includegraphics[clip=true, trim = 135 295 120 280, width=.32\linewidth]{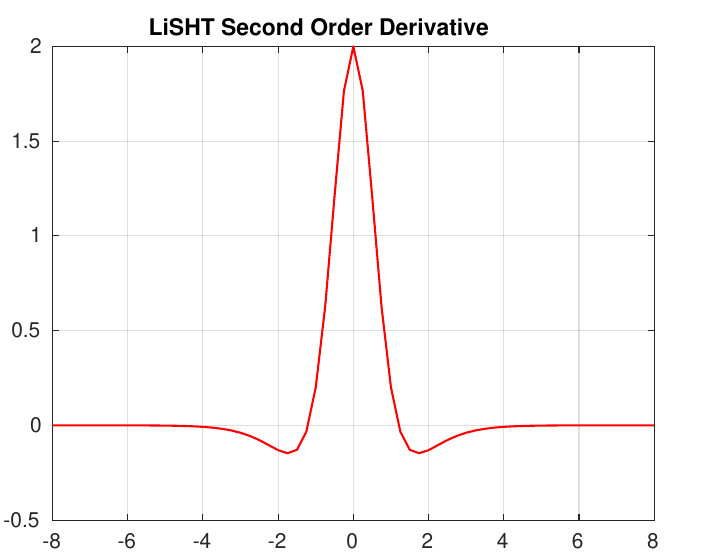} 
\\
(a) & (b) & (c)\\
\end{tabular}
\caption{(a) The characteristics of the \textit{LiSHT} activation function along with $Tanh$, $Sigmoid$, $ReLU$, and $Swish$. (b) The $1^{st}$ order Derivative of proposed \textit{LiSHT} activation. (c) The $2^{nd}$ order Derivative of the proposed \textit{LiSHT} activation.}
\label{fig:prop_derivative}
\end{figure*}

The contributions of this paper are as follows,
\begin{itemize}
\item A new activation function named non-parametric Linearly Scaled Hyperbolic Tangent ($LiSHT$) is proposed by linearly scaling the $Tanh$ activation function.
\item The increased amount of non-linearity of the proposed activation function is visualized from its first and second order derivatives curves (Fig.~\ref{fig:prop_derivative}).
\item The proposed $LiSHT$ activation function is tested with different types of neural networks, including Multilayer Perceptron, Residual Neural Network, and Long-Short Term Memory based networks.
\item Three different types of experimental data are used 1) $\mathbb{R}$-dimensional data, including Iris and MNIST (converted from image) datasets, 2) image data, including MNIST, CIFAR-10 and CIFAR-100 datasets, and 3) sentiment analysis data, including twitter140 dataset.
\item The impact of different non-linearity functions over activation feature maps and weight distribution has been analyzed.
\item The activation maps, weight distributions and optimization landscape are also analyzed to show the effectiveness of the proposed \textit{LiSHT} activation function.
\end{itemize}

This paper is organized as follows: Section 2 outlines the proposed $LiSHT$ activation; Section 3 presents the mathematical analysis; Section 4 describes the experimental setup; Section 5 presents the results; and Section 6 contains the concluding remarks.

\section{Proposed LiSHT Activation Function}
A Deep Neural Network~(DNN) comprises of multiple hidden nonlinear layer. Let an input vector be $x \in \mathbb{R}^{d}$, and each layer transforms the input vector followed by a nonlinear mapping from the $l^{th}$ layer to the $(l+1)^{th}$ layer as follows:  
\begin{equation} \label{equ:eq1}
\begin{split}
\begin{rcases}
\tau^{0} &= x \\
s_{i}^{l+1} &= \sum_{j=1}^{N^{l}}w_{ij}^{l}\tau_{j}^{l} + o_{i}^{l} \\
\tau_{i}^{l+1} &= \phi(s_{i}^{l+1})
\end{rcases}
\end{split}
\end{equation}

Here, $\tau$ represents the activation volume of any given layer, $s_{i}^{l}, w_{ij}^{l}, o_{i}^{l}$, and $N^{l}$ represent the vectors of output, weights, biases and number of units in the hidden $l^{th}$ layer, respectively, and a non-linear activation mapping $\phi(x)$. Looking for an efficient and powerful activation function in DNN is always demanding due to the overabundance by the \textit{saturation} properties of existing activation functions. An activation function $\phi(x)$ is said to be saturate~\cite{glorot2011deep}, if its derivative $\phi'(x)$ tends to zero in both directions (i.e., $x \rightarrow \infty$ and $x \rightarrow - \infty$, respectively). The training of a deep neural networks is almost impossible with of $Sigmoid$ and $Tanh$ activation functions due to the gradient diminishing problem when input is either too small or too large \cite{goodfellow2016deep}. For the first time, the Rectified Linear Unit ($ReLU$) (i.e., $\phi(x) = max(0,x)$) became very popular activation for training the DNN \cite{alexnet}. But, $ReLU$ also suffers due to the gradient diminishing problem for negative inputs which lead to the dying neuron problem.

Hence, we propose a non-parametric linearly scaled hyperbolic tangent activation function, so called $LiSHT$. Like $ReLU$ \cite{alexnet} and $Swish$~\cite{ramachandran2017swish}, $LiSHT$ shares the similar unbounded upper limits property on the right hand side of activation curve. However, because of the symmetry preserving property of $LiSHT$, the left hand side of the activation is in the upwardly unbounded direction, hence it satisfies non-monotonicity (see Fig.~\ref{fig:prop_derivative}(a)). Apart from the literature~\cite{clevert2015fast},\cite{ramachandran2017swish} and to the best of our knowledge, first time in the history of activation function, $LiSHT$ utilizes the benefits of positive valued activation without identically propagating all the inputs, which mitigates gradient vanishing at back propagation and acquiesces faster training of deep neural network. The proposed activation function $LiSHT$ is computed by multiplying the $Tanh$ function to its input $x$ and defined as, 
\begin{equation} \label{equ:eq2}
\begin{split}
\phi(x) &= x \cdot g(x) \\
\end{split}
\end{equation}
where $g(x)$ is a hyperbolic tangent function and defined as,
\begin{equation}
g(x) = Tanh(x) = \frac{exp^{x} - exp^{-x}}{exp^{x} + exp^{-x}}.
\end{equation}
where $x$ is the input to the activation function and $exp$ is the exponential function.

For the large positive inputs, the behavior of the $LiSHT$ is close to the $ReLU$ and $Swish$, i.e., the output is close to the input as depicted in Fig.~\ref{fig:prop_derivative}(a). Whereas, unlike $ReLU$ and other commonly used activation functions, the output of $LiSHT$ for negative inputs is symmetric to the output of $LiSHT$ for positive inputs as illustrated in Fig.~\ref{fig:prop_derivative}(a). 
The $1^{st}$ order derivative (i.e., $\phi'(x)$) of $LiSHT$ is given as follows,
\begin{equation} \label{equ:eq4}
\begin{split}
\phi'(x) &= x[1 - Tanh^{2}(x)] + Tanh(x) \\
        &= x + Tanh(x)[1 - \phi(x)].
\end{split}
\end{equation}
Similarly, the $2^{nd}$ order derivative (i.e., $\phi''(x)$) of $LiSHT$ is given as follows,
\begin{equation} \label{equ:eq5}
\begin{split}
\phi''(x) &= 1 - Tanh(x)\phi(x) + [1 - \phi(x)](1 - Tanh^2(x)) \\
      &= 2 - Tanh(x)\phi'(x) - \phi(x) - Tanh(z)[\phi'(x) - x] \\
      &= 2[1 - Tanh(x)\phi'(x)].
\end{split}
\end{equation}
The $1^{st}$ and $2^{nd}$ order derivatives of the proposed \textit{LiSHT} are plotted in Fig.~\ref{fig:prop_derivative}(b) and Fig.~\ref{fig:prop_derivative}(c), respectively. An attractive characteristic of the \textit{LiSHT} is \textit{self-stability} property, the magnitude of derivatives is less than $1$ for $x\in[-0.65, 0.65]$. It can be observed from the derivatives of $LiSHT$ in Fig.~\ref{fig:prop_derivative} that the amount of non-linearity is very high near to zero as compared to the existing activations which can boost the learning of a complex model. 
As described in Fig. \ref{fig:prop_derivative}(c) that the $2^{nd}$ order derivative of proposed $LisHT$ activation function is similar to the opposite of the Laplacian operator (i.e., the $2^{nd}$ order derivative of Gaussian operator) which is useful to maximize a function. Thus, due to opposite nature of Gaussian operator, the proposed $LiSHT$ activation function boosts the training of the neural network for the minimization problem of the loss function.

We understand that being unbounded for both negative and positive inputs, smooth, and non-monotonicity are the advantages of the proposed $LiSHT$ activation. The complete unbounded property makes $LiSHT$ different from all the traditional activation functions. Moreover, it makes use of strong advantage of positive feature space. $LiSHT$ is a smooth, symmetric w.r.t. y-axis and non-monotonic function and introduces more amount of non-linearity in the training process than $Swish$. 

\begin{figure}[!t]
\centering
\includegraphics[clip=true, trim = 05 335 50 05, width=0.7\columnwidth]{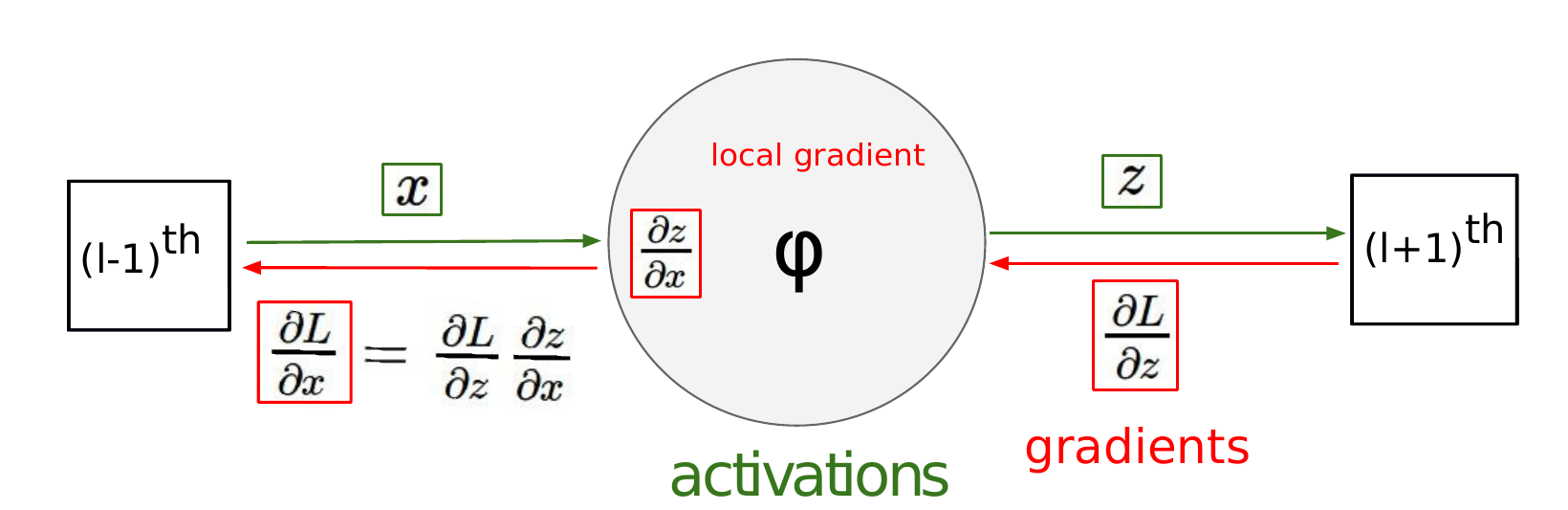}
\caption{Flow of gradients through any activation layer.}
\label{fig:analysis}
\end{figure}

\section{Mathematical Analysis}
In this section we show mathematically that $LiSHT$ actively solves the vanishing gradient problem of $Tanh$. The flow of gradient through any activation function is depicted in Fig. \ref{fig:analysis}. Let $\phi$ is an activation function given as $z = \phi(x)$, where $x$ is the input and $z$ is the output. Let $L$ is the final objective function and the running gradient $\frac{\partial L}{\partial z}$ is the input to $\phi$ during back-propagation. The running gradient output of $\phi$ is $\frac{\partial L}{\partial x} = \frac{\partial L}{\partial z}.\frac{\partial z}{\partial x}$ using chain rule, where $\frac{\partial z}{\partial x}$ is the local gradient of $\phi$. 
\begin{theorem}
If $\phi = Tanh$ then it leads to the gradient diminishing problem.
\end{theorem}
\begin{proof}
\begin{equation} \label{equ:eq6_1}
z = Tanh(x)
\end{equation}
The local gradient for $Tanh$ activation is given as,
\begin{equation} \label{equ:eq7}
\begin{split}
\frac{\partial z}{\partial x} = 1-Tanh^2(x) \approx \begin{cases} 0, & ~x < -2\\ \frac{\partial z}{\partial x}, & -2 \leq x \leq 2 \\
0, & ~x > 2 \end{cases}
\end{split}
\end{equation}
where $Tanh^2(x) \approx 1$ for $-2 < x < 2$. It can be noticed that for smaller and larger inputs the local gradient $\frac{\partial z}{\partial x}$ of $Tanh$ is very close to zero which makes the running gradient $\frac{\partial L}{\partial x}$ also close to zero, thus leading to the gradient diminishing problem.
\end{proof}

\begin{theorem}
If $\phi = LiSHT$ then the local gradient $\frac{\partial z}{\partial x} = 0$ iff $x=0$. 
\end{theorem}
\begin{proof}
\begin{equation} \label{equ:eq8}
z = LiSHT(x) = x.Tanh(x)
\end{equation}
The local gradient for LiSHT activation is given as,
\begin{equation} \label{equ:eq9}
\begin{split}
\frac{dz}{dx} = x + Tanh(x)[1-xTanh(x)] 
\end{split}
\end{equation}
For $x < -2$, $Tanh(x) \approx -1$, thus $\frac{\partial z}{\partial x} \approx -1$. For $x > 2$, $Tanh(x) \approx 1$, thus $\frac{\partial z}{\partial x} \approx 1$. For $-2 \leq x \leq 2$, $-1 \leq Tanh(x) \leq 1$, thus $2x-1 \leq \frac{\partial z}{\partial x} \leq 2x+1$. The $LiSHT$ can lead to gradient diminishing problem iff 
$\frac{\partial z}{\partial x} = x + Tanh(x)[1-xTanh(x)] = 0$. It means 
$x = \frac{Tanh(x)}{Tanh^2(x) - 1}$ which is only possible iff $x=0$. It can be also visualized in Fig. \ref{fig:prop_derivative}(b). 
Thus, the $LiSHT$ activation function exhibits non-zero gradient for all positive and negative inputs and solves the gradient diminishing problem of $Tanh$ activation function.
\end{proof}

\section{Experimental Setup}
In this section, first, six datasets are described in detail, then the three types of networks are summarized, and finally the training settings are stated in detail.

\subsection{Datasets Used}
We evaluate the proposed \textit{LiSHT} activation function on five benchmark databases, including Iris, MNIST, CIFAR-10, CIFAR-100 and twitter140. 
The \textbf{Fisher's Iris Flower dataset}\footnote{C. Blake, C. Merz, UCI Repository of Machine Learning Databases.}~\cite{zhang2014random} 
consists three Iris species (i.e., Versicolor, Virginica and Setosa) with a total of $150$ examples. Each example of Iris dataset is represented by four characteristics, including length and width of petal and sepal, respectively.
The \textbf{MNIST dataset} is a popular dataset to recognize the English digits (i.e., 0 to 9) in images. It consists of 60,000 and 10,000 images in the training and test sets, respectively \cite{mnist}. 
The \textbf{CIFAR-10 dataset} is an object recognition dataset with 10 categories having images of resolution $32\times32$ \cite{krizhevsky2009learning}. The 50,000 and 10,000 images are available in the training and test sets, respectively. 
All the images of the CIFAR-10 dataset are also present in the \textbf{CIFAR-100 dataset} dataset (i.e., $50K$ for training and $10K$ for testing), but categorized in 100 classes. The training and testing test sets contain $100$ classes in CIFAR-100 dataset. 
The \textbf{twitter140 dataset}~\cite{go2009twitter} is used to perform the classification of sentiments of Twitter messages by classifying as either positive, negative or neutral with respect to a query. In this dataset, we have considered 1,600,000 examples, where 85\% are used as training set and the rest 15\% as validation set. 

\begin{table}[!t]
\caption{The classification accuracy on Iris and MNIST datasets using different activation functions for MLP model.}
\centering
\begin{tabular}{p{0.14\columnwidth}|p{0.1\columnwidth}|p{0.09\columnwidth}|p{0.12\columnwidth}|p{0.13\columnwidth}|p{0.13\columnwidth}|p{0.13\columnwidth}|p{0.095\columnwidth}}
\hline
\multirow{2}{*}{Dataset} & \multicolumn{7}{c}{Activation Functions} \\ \cline{2-8} 
& Sigmoid & Tanh & ReLU \cite{alexnet} & PReLU \cite{prelu} & LReLU \cite{lrelu} & Swish \cite{ramachandran2017swish} & LiSHT\\ \hline
Iris & 96.23 & 96.26 & 96.41 & 97.11 & 96.53 & 96.34 & \textbf{97.33}\\ \hline
MNIST & 98.43 & 98.26 & 98.48 & 98.34 & 97.69 & 98.45 & \textbf{98.60}\\ \hline
\end{tabular}
\label{tab:plain}
\end{table}

\subsection{Tested Neural Networks}
We use three models, including a Multi-layer Perceptron (MLP), a widely used Pre-activated Residual Network (ResNet-PreAct)~\cite{he2016identity}), and a Long-Sort Term Memory (LSTM) to show the performance of activation functions. These architectures are explained in this section. 
The \textbf{Multi-layer Perceptron (MLP)} with one hidden layer is used in this paper for the classification of data. 
The internal architecture in MLP uses input, hidden and final $softmax$ layer with $6$, $5$, and $4$ nodes for the Car evaluation dataset. For Iris Flower dataset, the MLP uses $4$, $3$, and $3$ nodes in the input, hidden and final $softmax$ layer, respectively. The MNIST dataset samples are converted into 1-D vectors when used with MLP. Thus, for MNIST dataset, the MLP uses $784$ neurons in the input layer, $512$ neurons in the hidden layer, and $10$ neurons in the last layer.
The \textbf{Residual Neural Network (ResNet)} is a very popular CNN model for the image classification task. We use the Pre-activated ResNet \cite{he2016identity} for image classification experiments in this paper.
The ResNet-PreAct is used with 164-layer (i.e., very deep network) for CIFAR-10 and CIFAR-100 datasets, whereas it is used with 20-layer for MNIST dataset. 
The channel pixel mean subtraction is used for preprocessing over image datasets with this network as per the standard practice being followed by most image classification neural networks.
In this paper, the \textbf{Long Short Term Memory (LSTM)} is used as the third type of neural network, which basically belongs to the Recurrent Neural Network (RNN) family.
A single layered LSTM with $196$ cells is used for sentiment analysis over twitter140 dataset. The LSTM is fed with $300$ dimensional word vectors trained with FastText Embeddings. 

\begin{table}[!t]
\caption{The classification accuracy on MNIST and CIFAR-10/100 datasets using different activation functions for ResNet model.}
\centering
\begin{tabular}{p{0.14\columnwidth}|p{0.1\columnwidth}|p{0.09\columnwidth}|p{0.12\columnwidth}|p{0.13\columnwidth}|p{0.13\columnwidth}|p{0.13\columnwidth}|p{0.095\columnwidth}}
\hline
\multirow{2}{*}{Dataset} & \multirow{2}{*}{\begin{tabular}[c]{@{}c@{}}ResNet \\Depth\end{tabular}} & \multicolumn{6}{c}{Activation Functions} \\ \cline{3-8} 
& & Tanh & ReLU \cite{alexnet} & PReLU \cite{prelu} & LReLU \cite{lrelu} & Swish \cite{ramachandran2017swish} & LiSHT\\ \hline
MNIST & 20 & 99.48 & 99.56 & 99.56 & 99.52 & 99.53 & \textbf{99.59}\\ \hline
CIFAR-10 & 164 & 89.74 & 91.15 & 92.86 & 91.50 & 91.60 & \textbf{92.92}\\ \hline
CIFAR-100 & 164 & 68.80 & 72.84 & 73.01 & 72.24 & 74.45 & \textbf{75.32}\\ \hline
\end{tabular}
\label{tab:resnet}
\end{table}

\subsection{Training Settings}
We perform the implementation using in the Keras deep learning framework. Different computer systems, including different GPUs (such as NVIDIA Titan X, Pascal 12GB GPU and NVIDIA Titan V 12GB GPU) are used at different stages of the experiments. 
The $Adam$ optimizer \cite{adam,dubey2019diffgrad} is used for the experiments in this paper. The batch size is set to $128$ for the training of the networks. 
The learning rate is initialized to $0.1$ and reduced by a factor of $0.1$ at $80^{th}$, $120^{th}$, $160^{th}$, and $180^{th}$ epochs during training. For LSTM, after $10625$ iteration on $128$ sized mini-batches, the learning rate is dropped by a factor of $0.5$ up to $212,500$ mini-batch iterations.

\begin{table}[!t]
\caption{The classification accuracy on twitter140 dataset using different activation functions for LSTM model.}
\centering
\begin{tabular}{p{0.2\columnwidth}|p{0.15\columnwidth}|p{0.15\columnwidth}|p{0.15\columnwidth}|p{0.15\columnwidth}|p{0.15\columnwidth}}
\midrule
\multirow{2}{*}{Dataset} & \multicolumn{5}{c}{Activation Functions} \\ 
\cline{2-6} 
&  Tanh & ReLU \cite{alexnet} & LReLU \cite{lrelu} & Swish \cite{ramachandran2017swish} & LiSHT\\ \hline
Twitter140 &   82.27 & 82.47 & 78.18 & 82.22 & \textbf{82.47} \\ \hline
\end{tabular}
\label{tab:lstm}
\end{table}

\section{Results and Analysis}
We investigate the performance and effectiveness of the proposed $LiSHT$ activation and compare with state-of-the-art activation functions such as $Tanh$, $ReLU$, and $Swish$.

\subsection{Experimental Results}
The results on Iris and MNIST datasets using MLP model are reported in Table \ref{tab:plain}. The \textit{categorical cross-entropy} loss is used to train the models for $200$ epochs. In order to run training smoothly in both the dataset, $80\%$ of samples were randomly chosen for training and remaining $20\%$ are used for validation. The proposed $LiSHT$ activation achieves outperforms the existing activation functions. The top accuracy on Iris and MNIST datasets are achieved by LiSHT as $97.33\%$ and $98.60\%$, respectively. 

Table \ref{tab:resnet} shows the validation accuracies on MNIST, CIFAR-10 and CIFAR-100 datasets for different activations with pre-activation ResNet. The depth of ResNet is $20$ for MNIST and $164$ for CIFAR datasets. We train the model for $200$ epochs using the cross-entropy objective function. It is observed that $LiSHT$ outperforms the other activation functions on MNIST, CIFAR-10 and CIFAR-100 datasets, and achieves $99.59\%$ and $92.92\%$, and $75.32\%$ accuracy, respectively. Moreover, a significant improvement has been shown by $LiSHT$ on CIFAR datasets. The unbounded, symmetric and more non-linear properties of the proposed $LiSHT$ activation function facilitates better and efficient training as compared to the other activation functions such as $Tanh$, $ReLU$ and $Swish$. The unbounded and symmetric nature of $LiSHT$ leads to the more exploration of weights and positive and negative gradients to tackle the gradient diminishing and exploding problems. 

The sentiment classification performance in terms of the validation accuracy is reported in Table \ref{tab:lstm} over twitter140 dataset with LSTM for different activations. It is observed that the performance of proposed $LiSHT$ activation function is better than $Tanh$ and $Swish$, whereas the same as $ReLU$. It points out one important observation that by considering the negative values as negative by $Swish$ degrades the performance because it leads the $Swish$ activation more towards the linear function as compared to the $ReLU$ activation.

\begin{figure*}[!t]
\centering
\includegraphics[clip=true, trim = 25 11 70 42, width=0.49\columnwidth]{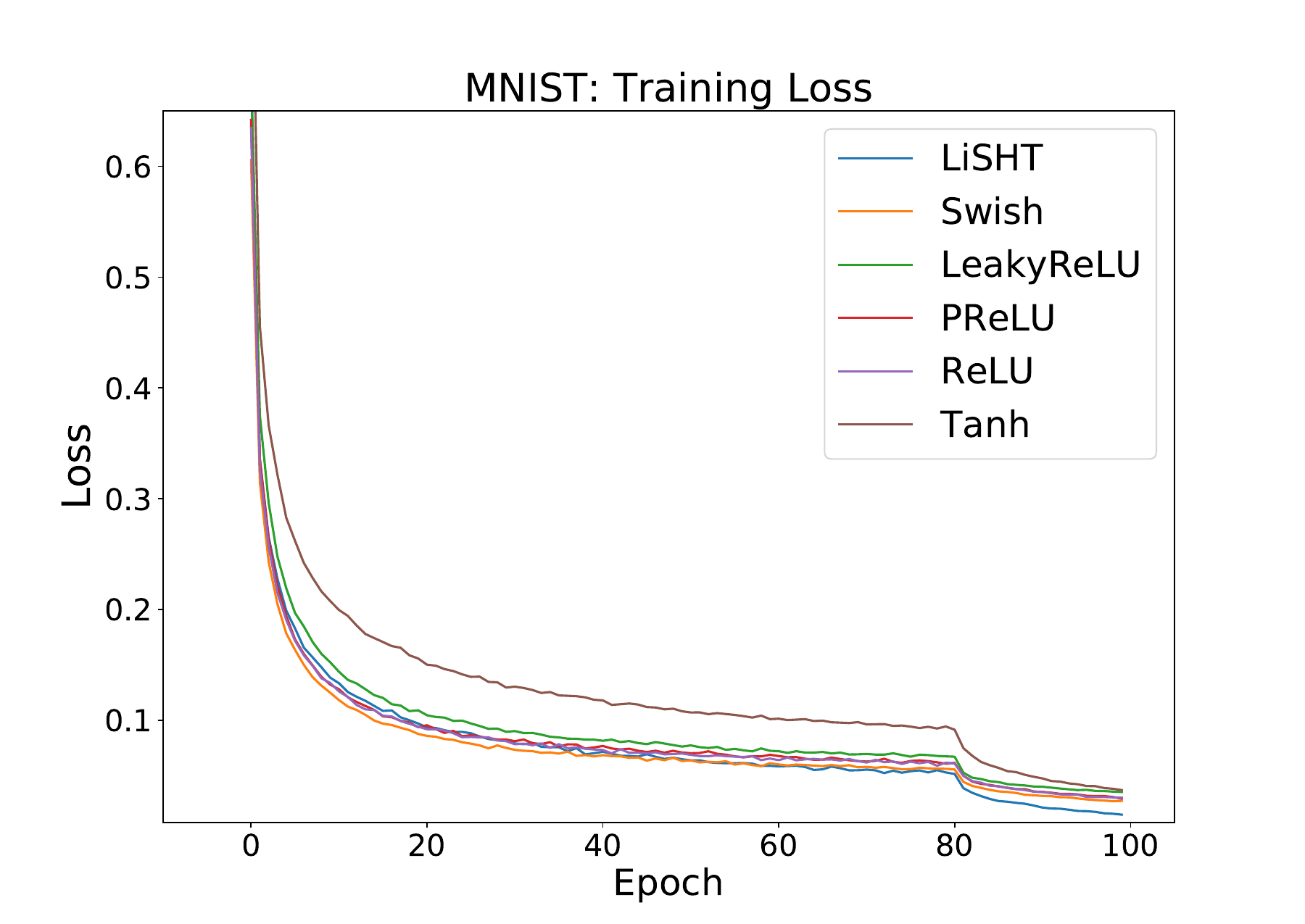}
\includegraphics[clip=true, trim = 40 11 70 42, width=0.49\columnwidth]{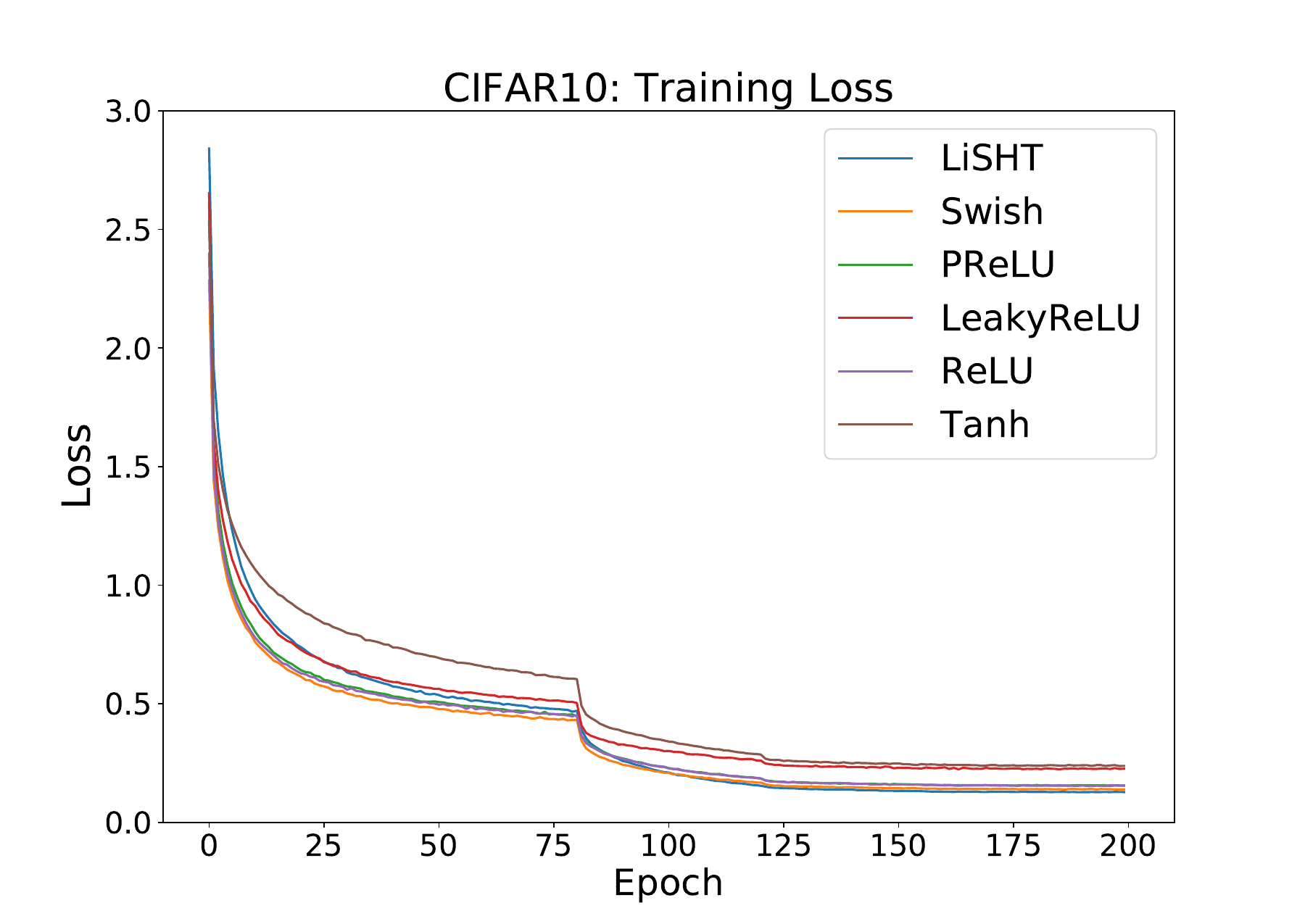}
\\
\includegraphics[clip=true, trim = 25 11 70 42, width=0.49\columnwidth]{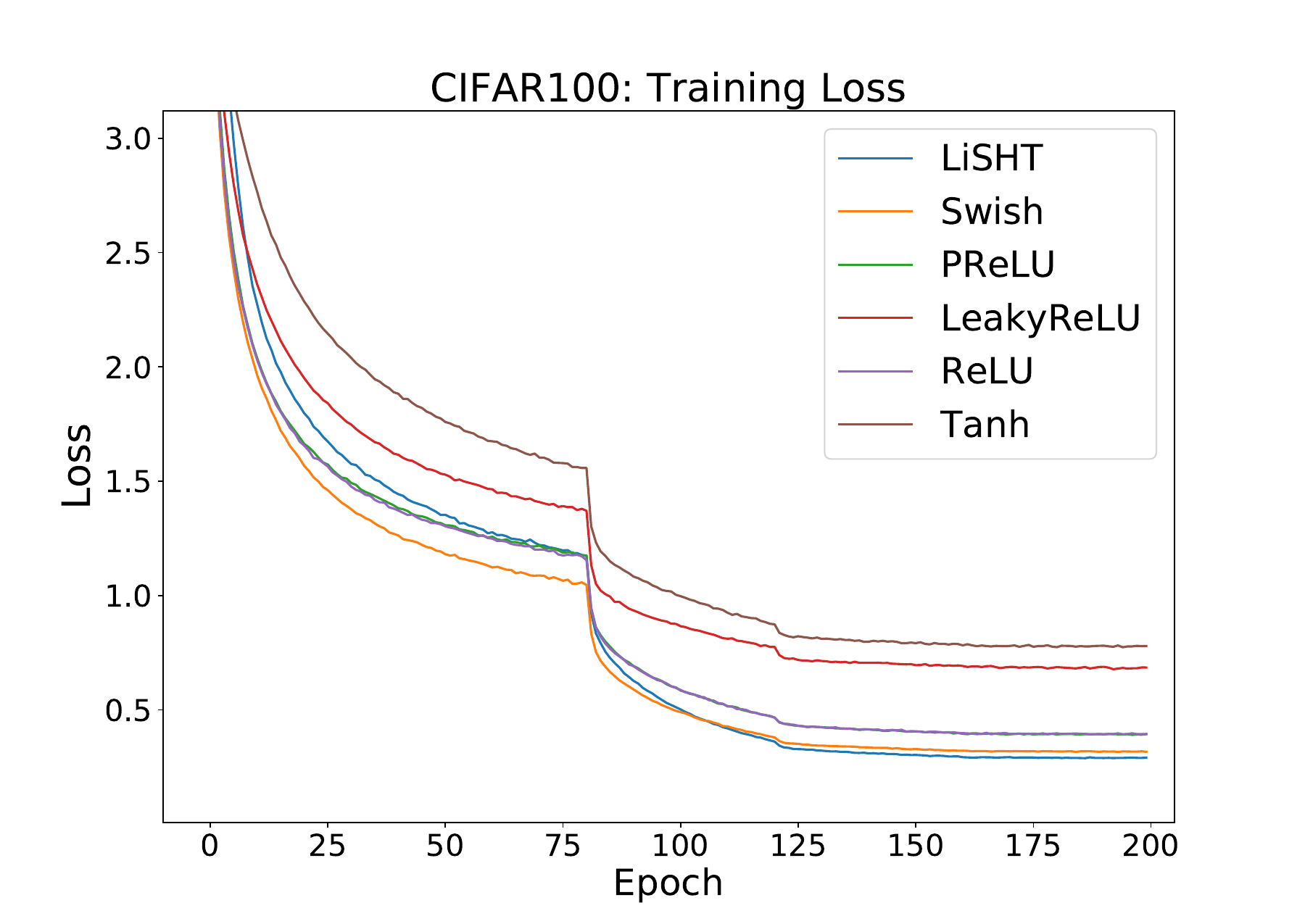}
\caption{The convergence curves in terms of loss for training sets using the $LiSHT$ and state-of-the-art $Tanh$, $ReLU$, and $Swish$ activations with ResNet model over MNIST (upper row, left column), CIFAR-10 (upper row, right column) and CIFAR-100 (lower row) datasets.}
\label{fig:error_rate_01}
\end{figure*}

\begin{figure*}[!t]
\centering
\includegraphics[clip=true, trim =45 11 60 42, width=0.49\columnwidth]{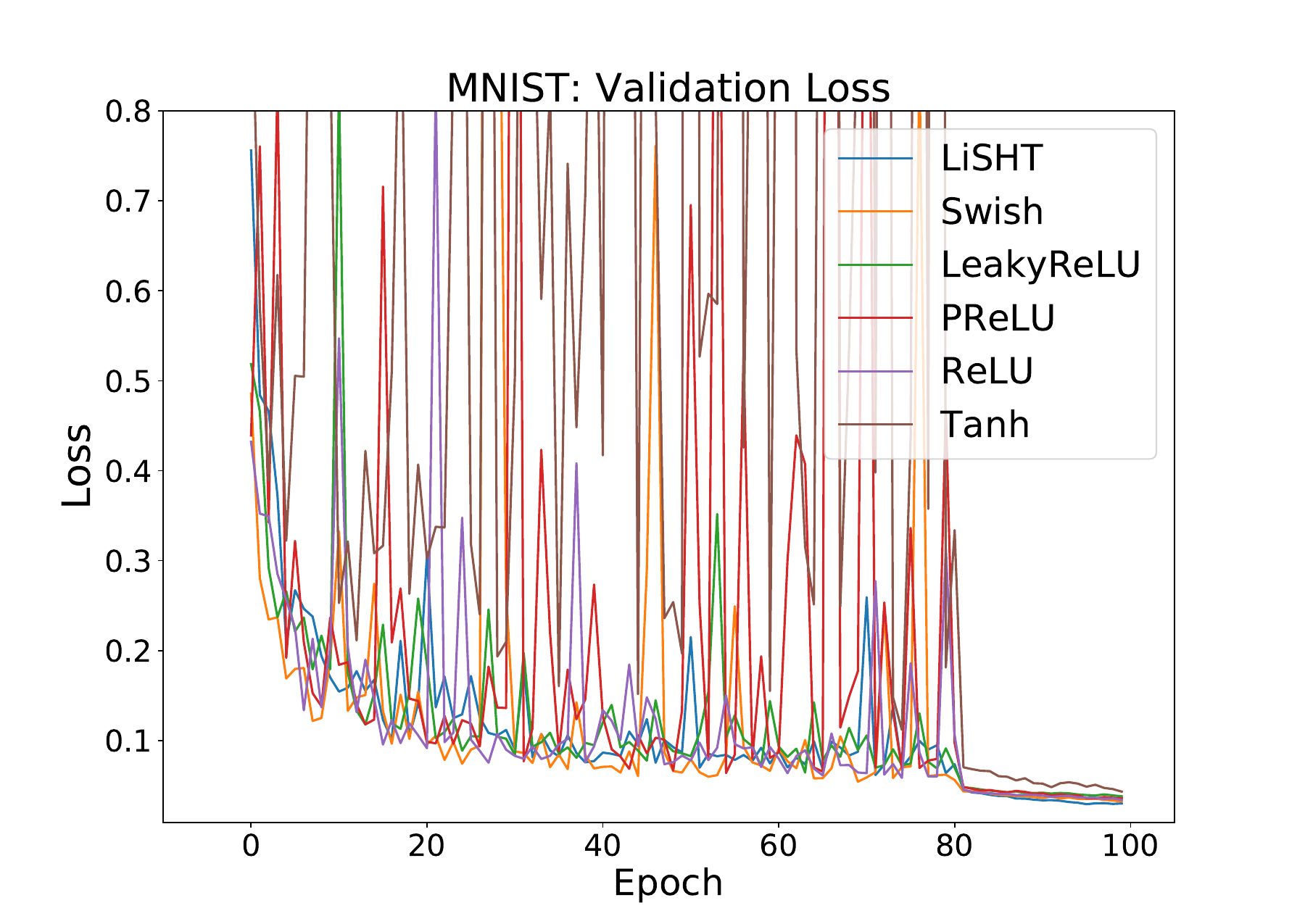}
\includegraphics[clip=true, trim = 40 11 70 42, width=0.49\columnwidth]{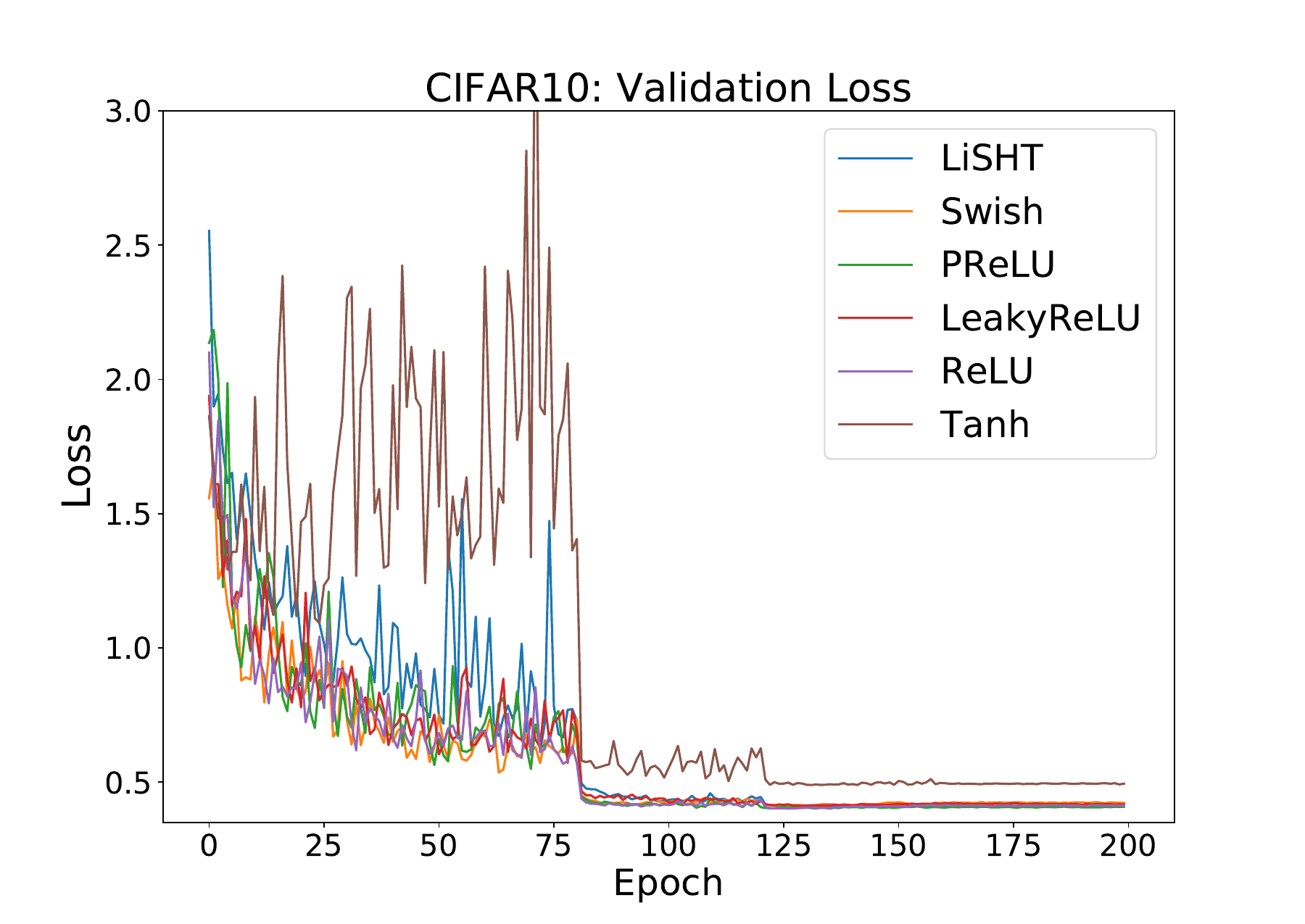}
\\
\includegraphics[clip=true, trim =45 11 60 42, width=0.49\columnwidth]{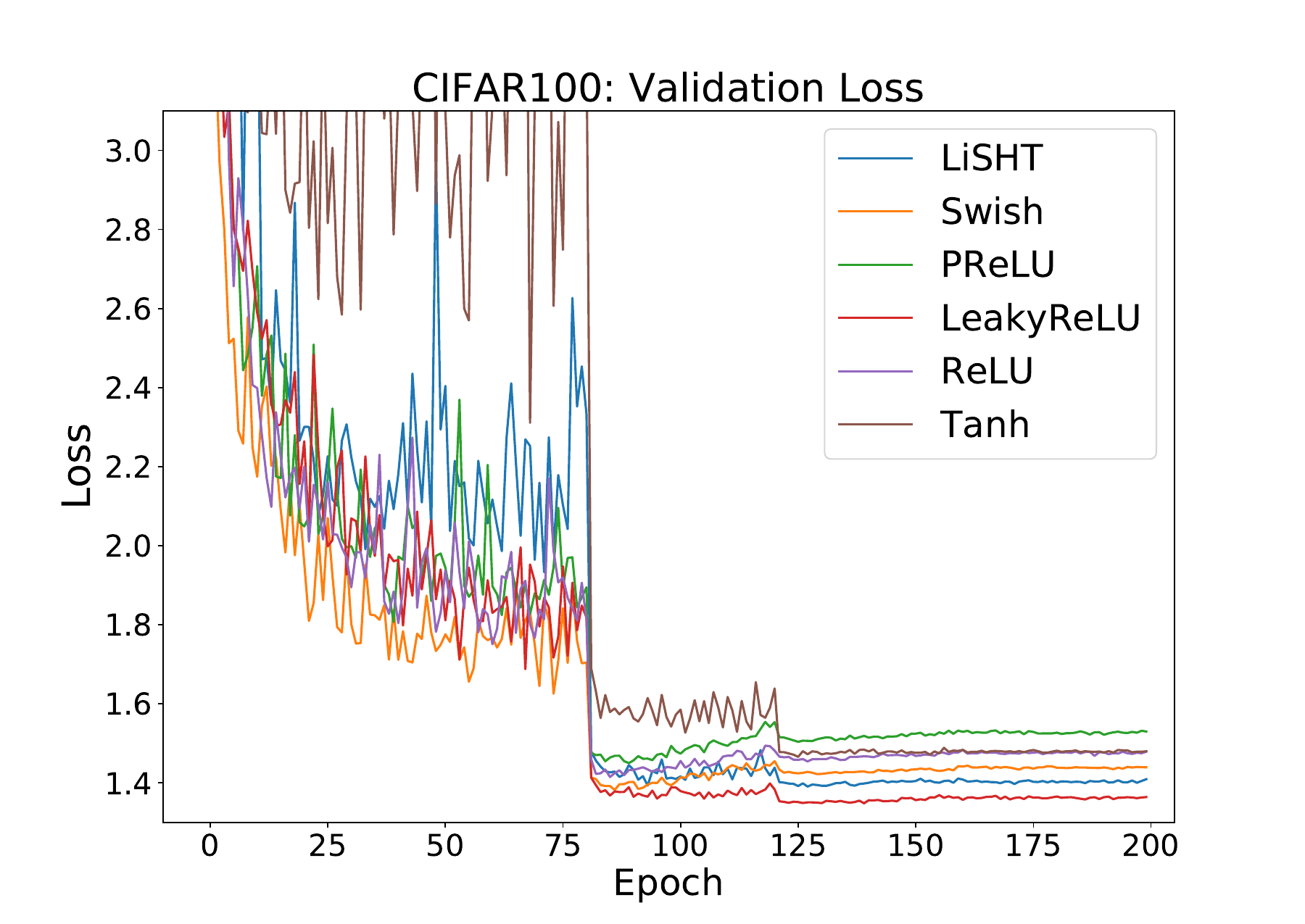}
\caption{The convergence curves in terms of loss for validation sets using the $LiSHT$ and state-of-the-art $Tanh$, $ReLU$, and $Swish$ activations with ResNet model over MNIST (upper row, left column), CIFAR-10 (upper row, right column) and CIFAR-100 (lower row) datasets.}
\label{fig:error_rate_02}
\end{figure*}

\begin{figure*}[!t]
\centering
\begin{tabular}{cc}
\includegraphics[clip=true, trim = 150 40 100 50, width=0.49\columnwidth, height=21mm]{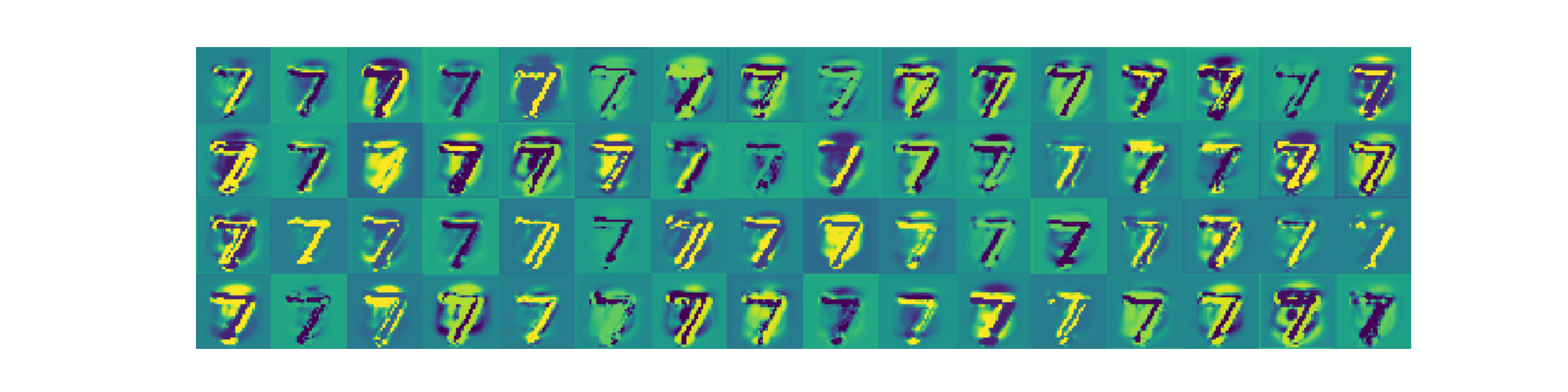}
&
\includegraphics[clip=true, trim = 150 40 100 50, width=0.49\columnwidth, height=21mm]{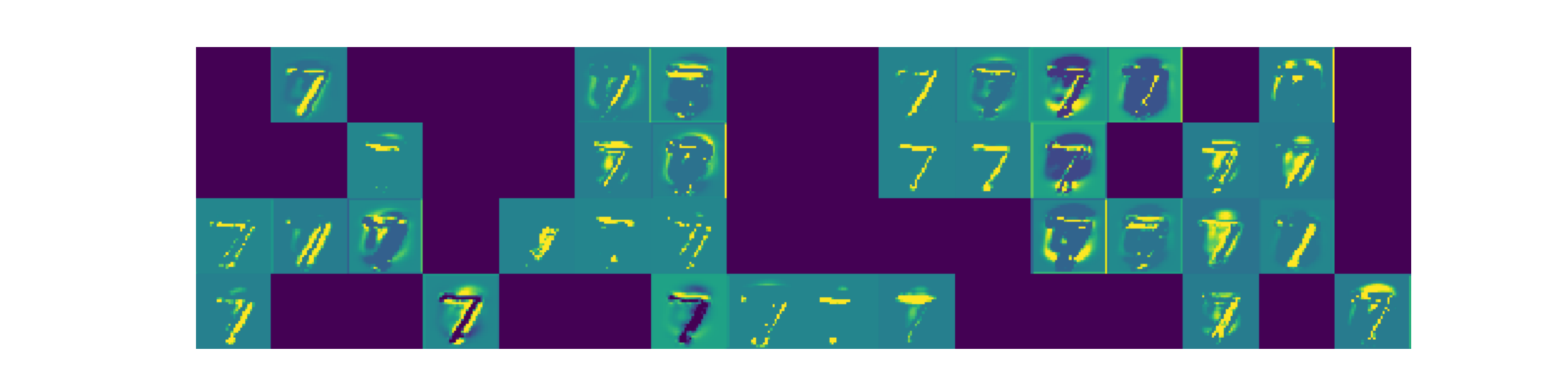}
\\
(a) Using $Tanh$ activation function & (b) Using $ReLU$ activation function \\
\includegraphics[clip=true, trim = 150 40 100 50, width=0.49\columnwidth, height=21mm]{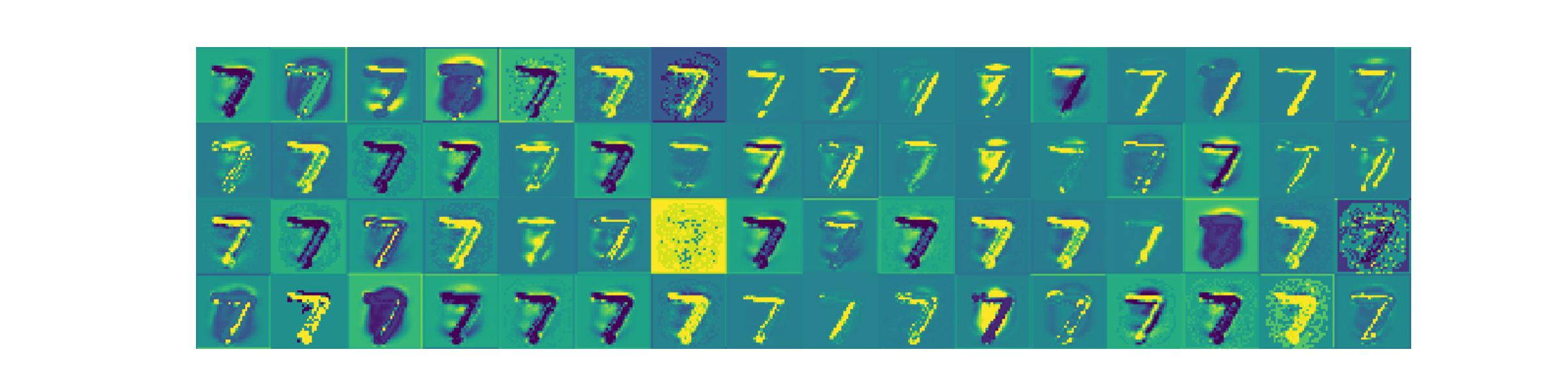}
&
\includegraphics[clip=true, trim = 150 40 100 50, width=0.49\columnwidth, height=21mm]{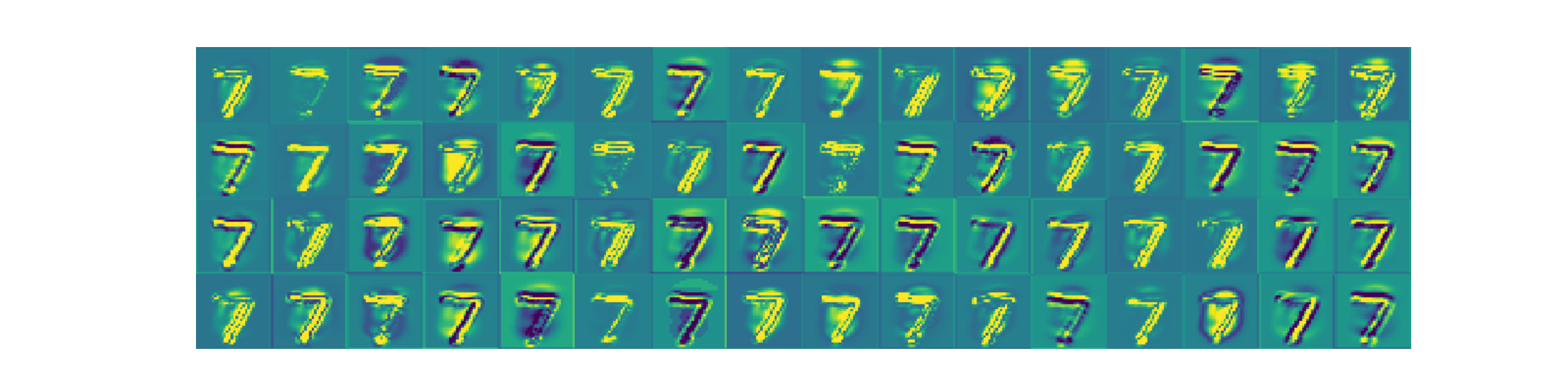}
\\
(c) Using $Swish$ activation function & (d) Using $LiSHT$ activation function \\
\end{tabular}
\caption{Visualization of MNIST digit $7$ from the $2^{nd}$ $conv$ layer activation feature maps without feature scale clipping using a fully trained pre-activation ResNet model using the (a) $Tanh$ (b) $ReLU$ (c) $Swish$ and (d) $LiSHT$ activation, respectively. Note that there are $64$ feature maps of dimension $32 \times 32$ in the $2^{nd}$ layer, represented in $4$ rows and $16$ columns.}
\label{fig:act_maps1}
\end{figure*}

\begin{figure*}[!t]
\centering
\begin{tabular}{cc}
\includegraphics[clip=true, trim = 20 10 40 40, width=0.49\columnwidth, height=25mm]{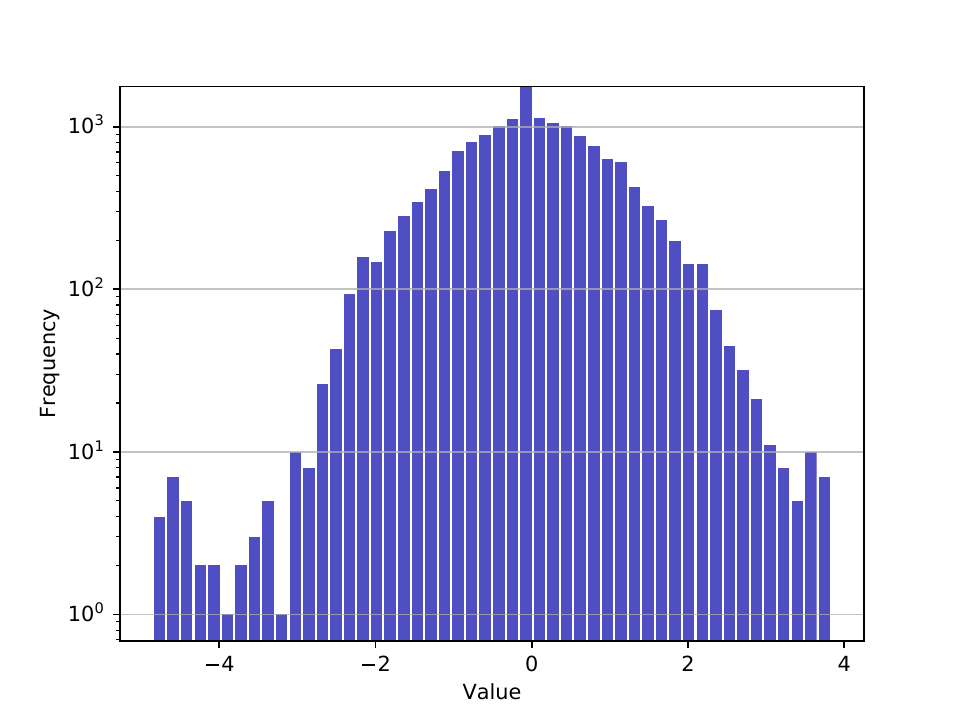}
&
\includegraphics[clip=true, trim = 18 10 20 40, width=0.49\columnwidth, height=25mm]{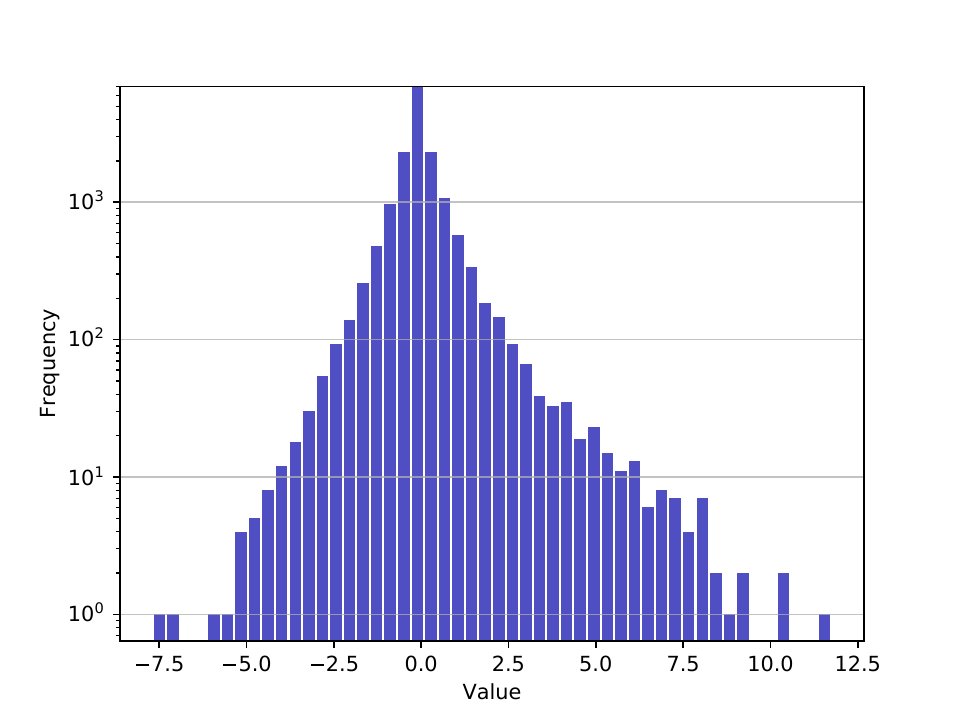}
\\
(a) & (b)\\
\includegraphics[clip=true, trim = 20 10 40 40, width=0.49\columnwidth, height=25mm]{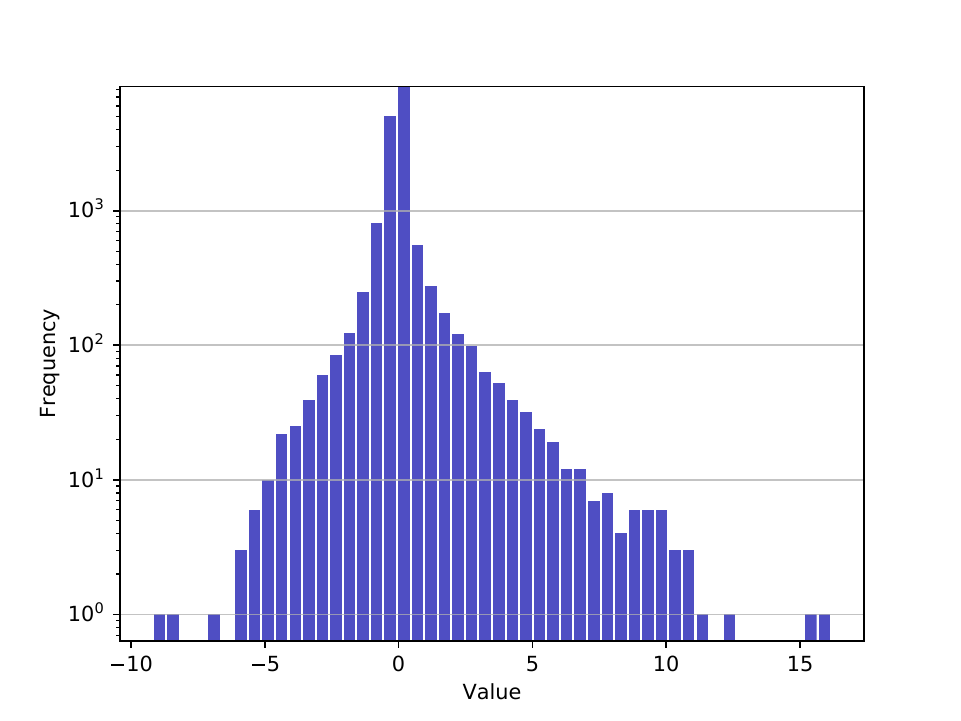}
&
\includegraphics[clip=true, trim = 18 10 20 40, width=0.49\columnwidth, height=25mm]{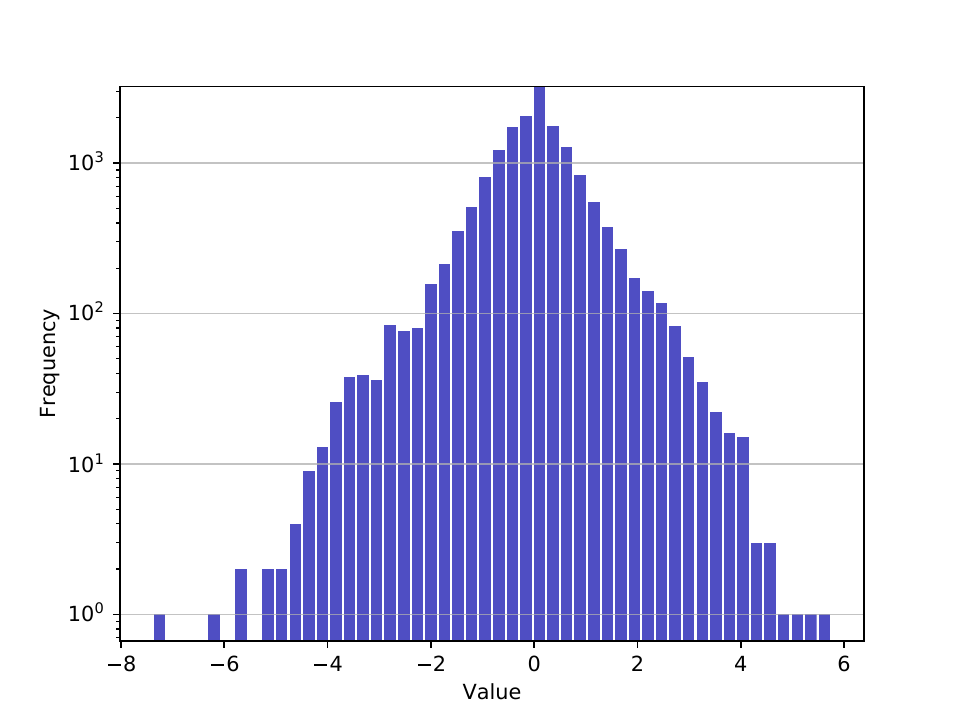}
\\
(c) & (d)
\end{tabular}
\caption{Visualizations of the distribution of weights from the final $Conv$ layer of pre-activation ResNet over MNIST dataset for the (a) $Tanh$ (b) $ReLU$ (c) $Swish$ and (d) $LiSHT$ activations, respectively.}
\label{fig:act_weight}
\end{figure*}

\subsection{Result Analysis}
The convergence curve of losses is also used as the metric to measure the learning ability of the ResNet model with different activation functions. The training and validation loss over the epochs are plotted in Fig. \ref{fig:error_rate_01} and \ref{fig:error_rate_02} for MNIST, CIFAR-10 and CIFAR-100 datasets using ResNet. 
It is clearly observed that the proposed $LiSHT$ boosts the convergence speed. 
It is also observed that the $LiSHT$ outperforms the existing non-linearities across several classification tasks with MLP, ResNet and LSTM networks. 

\subsection{Analysis of Activation Feature Maps}
In deep learning, it is a common practice to visualize the activations of different layer of the network.  In order to understand the effect of activation functions over the learning of important features at different layer, we have shown the activation feature maps for different non-linearities at $2^{nd}$ layer of the pre-activation ResNet of MNIST digit $7$ in Fig.~\ref{fig:act_maps1}. 
The number of activation feature maps in $2^{nd}$ and $11^{th}$ layers are $64$ (each having the $32 \times 32$ spatial dimensions) and $128$ (each having the $16 \times 16$ spatial dimensions), respectively. 
It can be seen from Fig.~\ref{fig:act_maps1} that the images looking deeper blue are due to the dying neuron problem caused by the non-learnable behavior arose due to the improper handling of negative values by the activation functions. The proposed $LiSHT$ activation consistently outperforms other activations. It is observed that the $LiSHT$ generates the less number of non-learnable filters due to the unbounded nature in both positive and negative scenarios which helps it to overcome from the problem of dying gradient. It is also observed that some image patches contain noise in terms of the Yellow color. The patches corresponding to the $LiSHT$ contain less noise. Moreover, it is uniformly distributed over all the patches, when $LiSHT$ is used, compared to other activation functions. It may be also one of the factors that proposed $LiSHT$ outperforms other activations.

\subsection{Analysis of Final Weight Distribution}
The weights of the layers are useful to visualize because it gives the idea about the learning pattern of the network in terms of 1) the positive and negative biasedness and 2) the exploration of weights caused by the activation functions. We have portrayed the weight distribution of final $Conv$ layer in Fig. \ref{fig:act_weight} for pre-activation ResNet over the MNIST dataset using $Tanh$, $ReLU$, $Swish$ and $LiSHT$ activations. The weight distribution for $Tanh$ is limited in between $-5$ and $4$ (see \ref{fig:act_weight}(a)) due to its bounded nature in both negative and positive regions. Interestingly, as depicted in \ref{fig:act_weight}(b), the weight distribution for $ReLU$ is biased towards the positive region because it converts all negative values to zero which restricts the learning of weights in the negative direction. This leads to the problems of dying gradient as well as gradient exploding. The $Swish$ tries to overcome the problems of $ReLU$, but unable to succeed due to the bounded nature in negative region (see \ref{fig:act_weight}(c)). The above mentioned problems are removed in the $LiSHT$ as suggested by its weight distribution shown in Fig. \ref{fig:act_weight}(d). The $LiSHT$ activation leads to the symmetric and smoother weight distribution. Moreover, it also allows the exploration of weights in the higher range (i.e., in between $-8$ and $6$ in the example of Fig. \ref{fig:act_weight}).

\begin{figure*}[!t]
\centering
\begin{tabular}{ccc}
\includegraphics[clip=true, trim = 0 0 0 0, width=0.32\columnwidth]{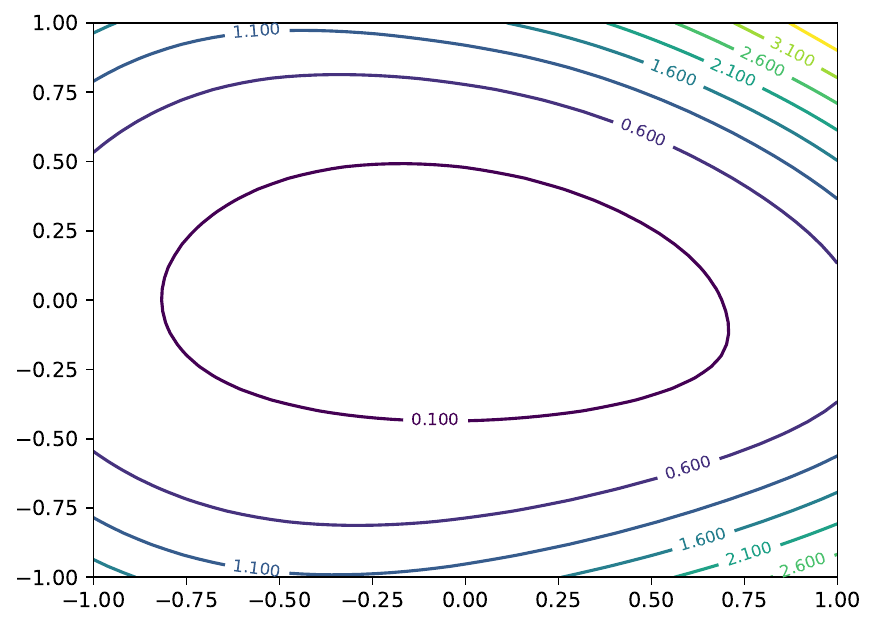}
&
\includegraphics[clip=true, trim = 0 0 0 0, width=0.32\columnwidth]{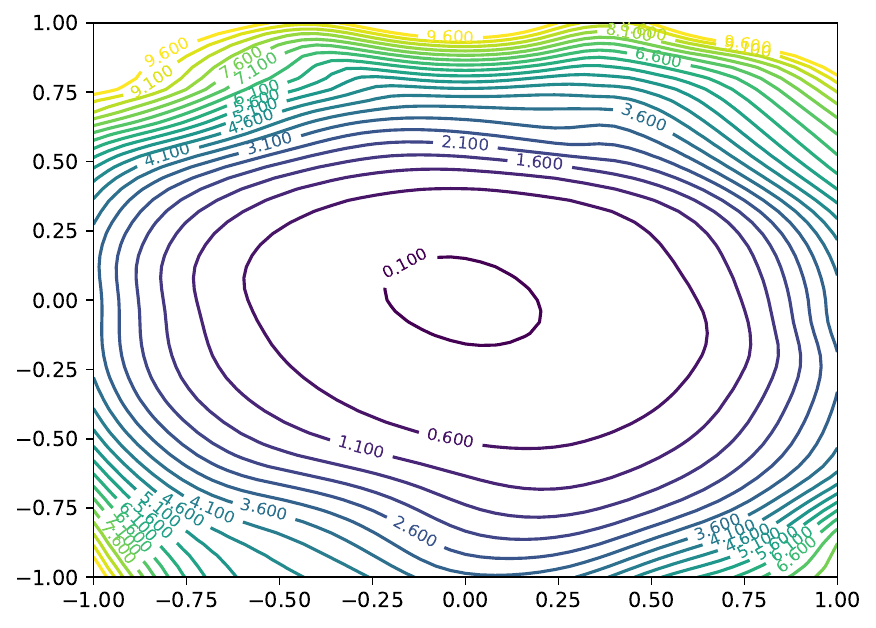}
&
\includegraphics[clip=true, trim = 0 0 0 0, width=0.32\columnwidth]{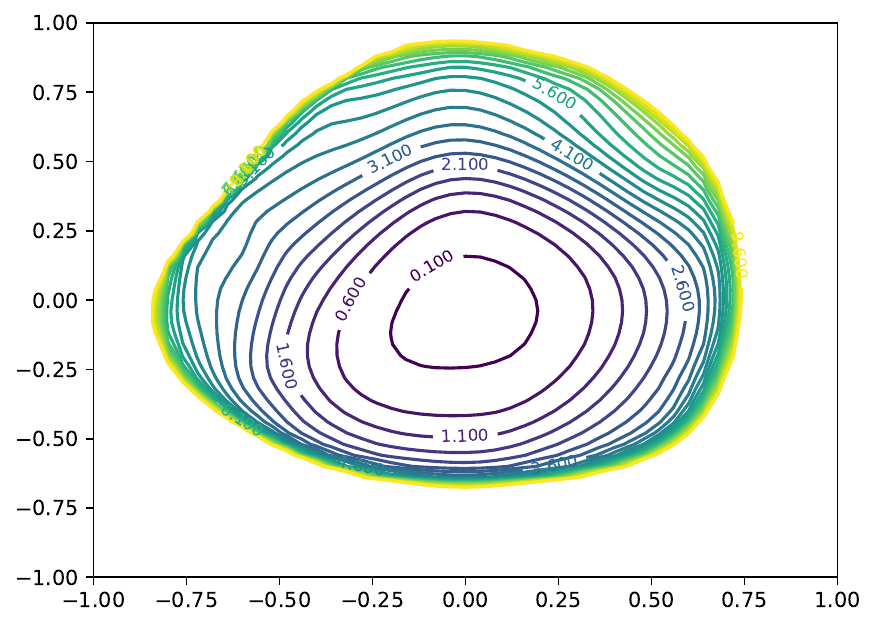}
\\
(a) ReLU & (b) Swish & (c) LiSHT
\end{tabular}
\caption{The visualization of 2D Loss Landscape plot of CIFAR-10 shown using ReLU, Swish and LiSHT, respectively.}
\label{fig:loss_land_01}
\end{figure*}

\begin{figure*}[!t]
\centering
\begin{tabular}{ccc}
\includegraphics[width=0.32\columnwidth, height=3.0cm]{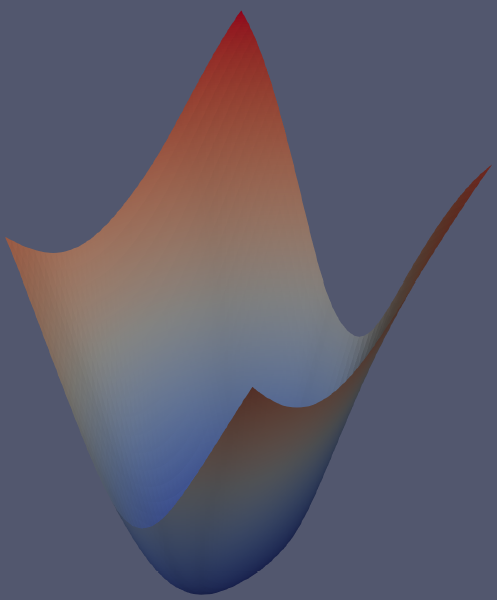}
&
\includegraphics[width=0.32\columnwidth, height=3.0cm]{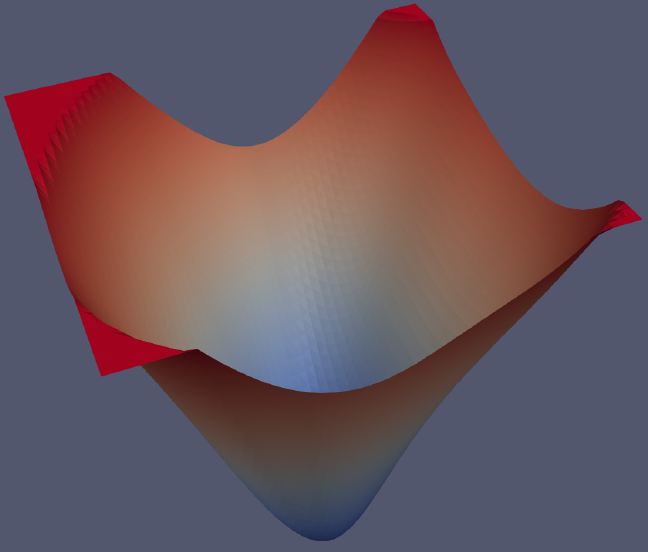}
&
\includegraphics[width=0.32\columnwidth, height=3.0cm]{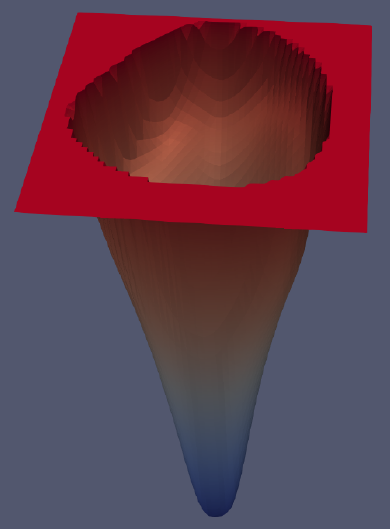}
\\
(a) ReLU & (b) Swish & (c) LiSHT
\end{tabular}
\caption{The visualization of 3D Loss Landscape plot of CIFAR-10 shown using ReLU, Swish and LiSHT, respectively.}
\label{fig:loss_land_02}
\end{figure*}

\subsection{Analysis of Loss Landscape}
The training ability of DNN is directly and indirectly influenced by the factors like network architecture, the choice of optimizer, variable initialization, and most importantly, what kind of non-linearity function to be used in the architecture. In order to understand the effects of network architecture on non-convexity, we trained the ResNet-152 using $ReLU$, $Swish$ and proposed $LiSHT$ activations and try to explore the structure of the neural network loss landscape. The $2D$ and $3D$ visualizations of loss landscapes are illustrated in Fig.~\ref{fig:loss_land_01} and ~\ref{fig:loss_land_02} by following the visualization technique proposed by Li et al. \cite{li2018visualizing}.

As depicted in the $2D$ loss landscape visualizations in Fig.~\ref{fig:loss_land_01}(a)-(c), the $LiSHT$ makes the network to produce the smoother loss landscapes with smaller convergence steps which is populated by the narrow, and convex regions. It directly impacts the loss landscape. However, $Swish$ and $ReLU$ also produce smooth loss landscape with large convergence steps, but unlike $LiSHT$, both $Swish$ and $ReLU$ cover wider searching area which leads to poor training behavior. In $3D$ landscape visualization, it can be seen in Fig.~\ref{fig:loss_land_02}(a)-(c), it can be observed that the slope of the $LiSHT$ loss landscape is higher than the $Swish$ and $ReLU$ which enables to train deep network efficiently. Therefore, we can say that, the $LiSHT$ decreases the non-convex nature of overall loss minimization landscape as compared to the $ReLU$ and $Swish$ activation functions.

\section{Conclusion}
A novel non-parametric linearly scaled hyper tangent activation function ($LiSHT$) is proposed in this paper for training the neural networks.
The proposed $LiSHT$ activation function introduces more non-linearity in the network. It is completely unbounded and solves the problems of diminishing gradient problems. Other properties of $LiSHT$ are symmetry, smoothness and non-monotonicity, which play an important roles in training.
The classification results are compared with the state-of-the-art activation functions. The efficacy of LiSHT is tested on benchmark datasets using MLP, ResNet and LSTM models. The experimental results confirm the effectiveness of the unbounded, symmetric and highly non-linear nature of the proposed $LiSHT$ activation function. The importance of unbounded and symmetric non-linearity in both positive and negative regions are analyzed in terms of the activation maps and weight distribution of the learned network. The visualization of loss landscape confirms the effectiveness of the proposed activations to make the training more smoother with faster convergence.

\section*{Acknowledgments}
We gratefully acknowledge the support of NVIDIA Corporation with the donation of the GeForce Titan X Pascal GPU used partially in this research.

\bibliographystyle{splncs04}
\bibliography{Reference}

\end{document}